\documentclass[10pt,twocolumn,letterpaper]{article}

\usepackage{cvpr}              % To produce the CAMERA-READY version
\usepackage[dvipsnames]{xcolor}

\definecolor{cvprblue}{rgb}{0.21,0.49,0.74}
\usepackage[pagebackref,breaklinks,colorlinks,citecolor=cvprblue]{hyperref}
\usepackage{float}
\usepackage{textcomp}
\usepackage[T1]{fontenc}
\usepackage{multirow}
\usepackage{stfloats}
\usepackage{wrapfig}
\usepackage{multicol}
\usepackage{float}

\newcommand{\gpt}{GPT-4V\xspace}
\newcommand{\dalle}{DALL$\cdot$E\xspace2\xspace}

\def\paperID{2} % *** Enter the Paper ID here
\def\confName{CVPR}
\def\confYear{2024}

\title{GPT as Psychologist? Preliminary Evaluations for GPT-4V on Visual Affective Computing}

\author{Hao LU$^{1,2,}$\thanks{Equal contribution.}$^{\ , }$\thanks{Project Leader.}$\ $, Xuesong NIU$^{3,*,\dag}$, Jiyao WANG$^{1,2,*}$, Yin WANG$^{4,*}$, Qingyong HU$^{2,*}$, Jiaqi TANG$^{1,2,*}$, \\
Yuting ZHANG$^{1,2}$, Kaishen YUAN$^{5}$, Bin HUANG$^{6}$, Zitong YU$^{5,}$\thanks{Corresponding author.}$\ $, Dengbo HE$^{1,2}$,
Shuiguang DENG$^{4}$,\\ Hao CHEN$^{2}$, Yingcong CHEN$^{1,2,\ddag}$, Shiguang SHAN$^{7}$\\
$^{1}$The Hong Kong University of Science \& Technology (Guangzhou), \\
$^{2}$The Hong Kong University of Science \& Technology, 
$^{3}$Beijing Institute for General Artificial Intelligence, \\
$^{4}$Zhejiang University,  $^{5}$Great Bay University, $^{6}$Hangzhou Research Institute, Beihang University,\\ $^{7}$Institute of Computing Technology, Chinese Academy of Sciences
}

\begin{document}
\maketitle
\begin{abstract}

% Multimodal language models are designed to process and integrate information from multiple modes, such as text, speech, images, and video. Using GPT4 to evaluate the performance of multimodal language models is critical to the development of the field. This paper assesses the application of multimodal language models in affective computing, especially in the ability to recognize facial action units, expressions, micro-expressions, micro-gestures, and non-contact physiological measurement tasks. The results show that GPT4 has high accuracy in recognizing facial action units and detecting micro-expressions. However, its performance in general facial expression recognition is not accurate. The paper also highlights the challenges of achieving high-precision micro-expression recognition and the potential for continued research and development in this area. In addition, the paper demonstrates the versatility and potential of GPT4 for handling advanced tasks in emotion recognition and related fields by integrating with task-related agents for more complex tasks, such as estimating heart rate through signal processing. In conclusion, this paper provides valuable insights into the potential applications and challenges of multimodal language models in emotion recognition and related fields.

Multimodal large language models (MLLMs) are designed to process and integrate information from multiple sources, such as text, speech, images, and videos. Despite its success in language understanding, it is critical to evaluate the performance of downstream tasks for better human-centric applications. This paper assesses the application of MLLMs with 5 crucial abilities for affective computing, spanning from visual affective tasks and reasoning tasks. The results show that \gpt has high accuracy in facial action unit recognition and micro-expression detection while its general facial expression recognition performance is not accurate. We also highlight the challenges of achieving fine-grained micro-expression recognition and the potential for further study and demonstrate the versatility and potential of \gpt for handling advanced tasks in emotion recognition and related fields by integrating with task-related agents for more complex tasks, such as heart rate estimation through signal processing. In conclusion, this paper provides valuable insights into the potential applications and challenges of MLLMs in human-centric computing. Our interesting example is in \url{https://github.com/EnVision-Research/GPT4Affectivity}.

\end{abstract}    
\section{Introduction}
\vspace{-1mm}
\label{sec:intro}

The development of multimodal large language models (MLLMs) has been a topic of growing interest in recent years~\cite{lv2024video,li2023videochat,maaz2023video,ye2023mplug,luo2023valley}. MLLMs are designed to process and integrate information from multiple modalities, such as text, speech, images, and videos. The development of these models has been driven by the need to improve the accuracy and efficiency of various tasks, such as affective computing, sentiment analysis, and natural language understanding.

MLLMs have shown great promise in improving the accuracy and robustness of affective computing systems~\cite{wang2022systematic,poria2017review}. These models can process and integrate information from multiple modalities, such as facial expressions, speech patterns, and physiological signals, to infer emotional states accurately~\cite{li2023videochat,maaz2023video,ye2023mplug,luo2023valley} with significant implications for various applications, such as healthcare, education, and human-computer interaction.
\begin{figure}[t]
\begin{center}
\includegraphics[scale=0.12]{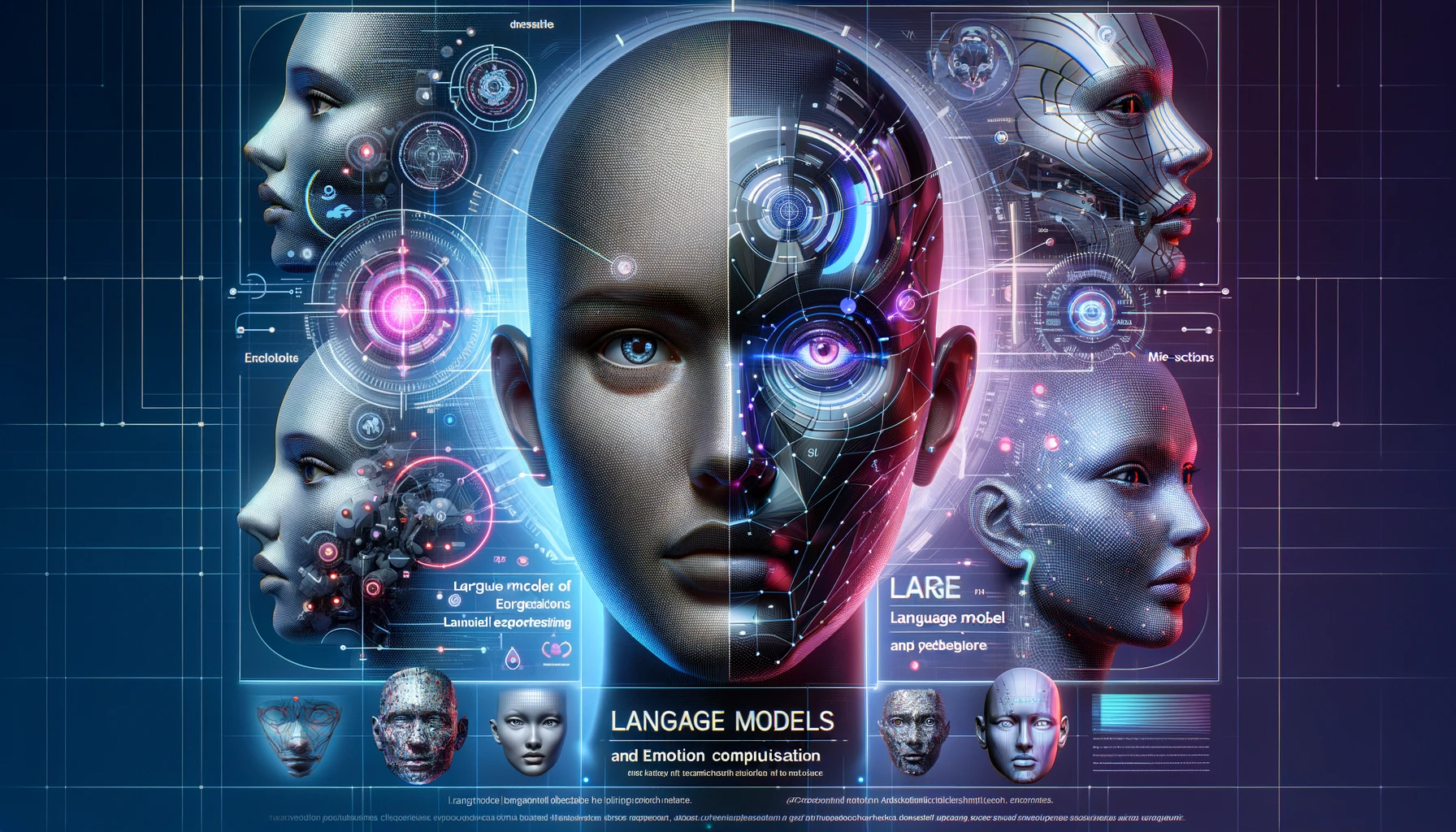}
\end{center}
\vspace{-4mm}
\caption{The propaganda image was generated by \dalle.}
\vspace{-8mm}
\label{fig:frame}
\end{figure}

Despite the rapid development of MLLMs, the need for standardized evaluation metrics is highlighted for accurate assessment. 
Different from the general language understanding evaluation benchmark that has been widely used to evaluate the performance of language models in NLP tasks~\cite{xu2020clue,wilie2020indonlu,wen2023road}, similar benchmarks to evaluate the performance of MLLMs for affective computing tasks are lacked, which is of great benefit to advance this field.

\gpt is the state-of-the-art MLLM that has shown remarkable success in various natural language processing tasks~\cite{achiam2023gpt}. Its ability to process and integrate information from multiple modalities makes it an ideal candidate for evaluating the performance of MLLMs in tasks for affective computing. Furthermore, it can invoke a variety of tools that benefit affective computing tasks, such as related program generation with self-correction. For example, \gpt can call \dalle to generate high-quality visual affective images shown in Fig.~\ref{fig:frame}.

In this paper, we evaluate \gpt with 5 typical human-centric tasks, spanning from visual affective tasks and reasoning tasks. We summarize our findings as follows: 

% The use of \gpt~\cite{yang2023dawn} in evaluating the performance of MLMs is crucial to advancing this field. \gpt is a state-of-the-art language model that has shown remarkable success in various natural language processing tasks~\cite{achiam2023gpt}. Its ability to process and integrate information from multiple modalities makes it an ideal candidate for evaluating the performance of MLMs in tasks such as affective computing and sentiment analysis. 
% In addition, large language models can help researchers study affective computing tasks by invoking a variety of tools. For example, in data processing and algorithm design, \gpt can write programs and try to run them by providing requirements. You can also offer some suggestions. \gpt can even help researchers build images by invoking some drawing tools. For example, \gpt can call Dell to generate high-quality visual affective images as shown in Fig.~\ref{fig:frame}.

% \gpt is one of the most powerful visual language model available. We have evaluated the capabilities of \gpt across multiple visual affective tasks, and our conclusions are as follows:

(1) \gpt is highly accurate in recognizing facial action units. This accuracy can be attributed to its advanced understanding of facial movements and their corresponding emotions, which allows it to effectively identify and analyze facial action units.

(2) \gpt is also precise in detecting micro-expressions. Its ability to process subtle and transient facial expressions enables it to accurately capture these fleeting emotional cues, which are often difficult for humans to perceive.

(3) \gpt's performance in general facial expression recognition is not as accurate. This limitation may be due to the complexity and variety of facial expressions, resulting in the challenges in capturing and analyzing them. Nevertheless, when \gpt is used to process thought chains, its accuracy in facial expression recognition improves significantly. This improvement suggests that incorporating additional contextual information is of great importance to recognize facial expressions.

(4) Achieving high accuracy in micro-expression recognition remains a challenging task. This difficulty arises from the transient nature of micro-expressions and the need to detect and classify them within a very short time period. These challenges call for continuing research and development in this area for improving affective computing
% Despite these challenges, continuing research and development in this area hold great potential for improving affective computing and understanding in various applications, such as mental health monitoring and virtual human companion systems.

(5) \gpt can also integrate with task-related agents to handle more complex tasks, such as detecting subtle facial changes and estimating heart rate with signal processing. By leveraging Python's powerful libraries and tools, \gpt can effectively process and analyze intricate facial data to derive valuable insights, such as heart rate estimation, which can further enhance its applications in mental health monitoring and virtual human companion systems.

% (5) \gpt can also integrate with task-related agents with Python to handle more complex tasks, such as detecting subtle facial changes and estimating heart rate through signal processing. This capability demonstrates \gpt's versatility and potential for handling advanced tasks in affective computing and related fields. By leveraging Python's powerful libraries and tools, \gpt can effectively process and analyze intricate facial data to derive valuable insights, such as heart rate estimation, which can further enhance its applications in mental health monitoring and virtual human companion systems.

\section{Visual Affective Evaluation}
Affective computing emerges as an interdisciplinary domain, leveraging computational technologies to discern, comprehend, and emulate human emotions. Its objective is to augment human-computer interaction, enhance user experiences, and facilitate improved communication and self-expression. Within the scope of computer vision, the analysis of human facial units~\cite{niu2019local}, expressions~\cite{shan2018reliable}, micro-expressions~\cite{zhao2023facial}, micro-gestures~\cite{liu2021imigue}, and deception detection~\cite{guo2023audio}, alongside physiological measurements~\cite{yu2021facial}, are pivotal to advancing emotional computing. Notably, large-scale pre-trained models, such as \gpt, have demonstrated substantial advancements in natural language processing, suggesting their considerable promise for application in affective computing. This study proposes to scrutinize the efficacy of \gpt across a variety of tasks, employing methodologies that include iterative conversations, open-ended inquiries, as well as multiple-choice and true/false questions.

\subsection{Action Unit Detection}
The Facial Action Coding System (FACS)~\cite{FACS} offers an explainable and reliable framework for the analysis of human facial expressions. It systematically deconstructs facial expressions into discrete components, known as Action Units (AUs), which correspond to the activation of specific facial muscles or groups thereof. Through the identification and quantification of these AUs, researchers can conduct a methodical examination of facial expressions and the emotional states they signify. Our assessment of \gpt's performance on the DISFA dataset~\cite{disfa}, utilizing a gamut of question types, underscores its proficiency in accurately identifying AUs, thereby enabling precise emotion recognition from minimal interaction.

Remarkably, \gpt exhibits exceptional accuracy in AU identification, facilitating nearly flawless judgment across all AUs examined as shown in Tab.~\ref{tab:au}. Although our presentation includes a limited number of examples, our comprehensive evaluation reveals \gpt's surprising efficacy in this domain. To quantitatively appraise this performance, we adopted a quantitative analysis approach, benchmarking against the F1 metrics as reported in related studies. 

Following~\cite{jacob2021facial}, we report F1 metrics on DISFA. Specifically, we judge whether the recognition is successful by searching whether there is AU$X$ (such as AU1) keyword in the reply question. The results show that the performance of \gpt is stronger than that of later professional models. This shows that \gpt has learned the micro-characteristics of emotion in a large number of network data and achieved significant recognition accuracy. Our findings indicate that \gpt's performance surpasses that of subsequent specialized models, underscoring its adeptness at learning the nuanced characteristics of emotion through extensive analysis of online data, thus achieving remarkable accuracy in emotion recognition.

\begin{figure*}[!ht]
\begin{center}
\includegraphics[scale=0.7]{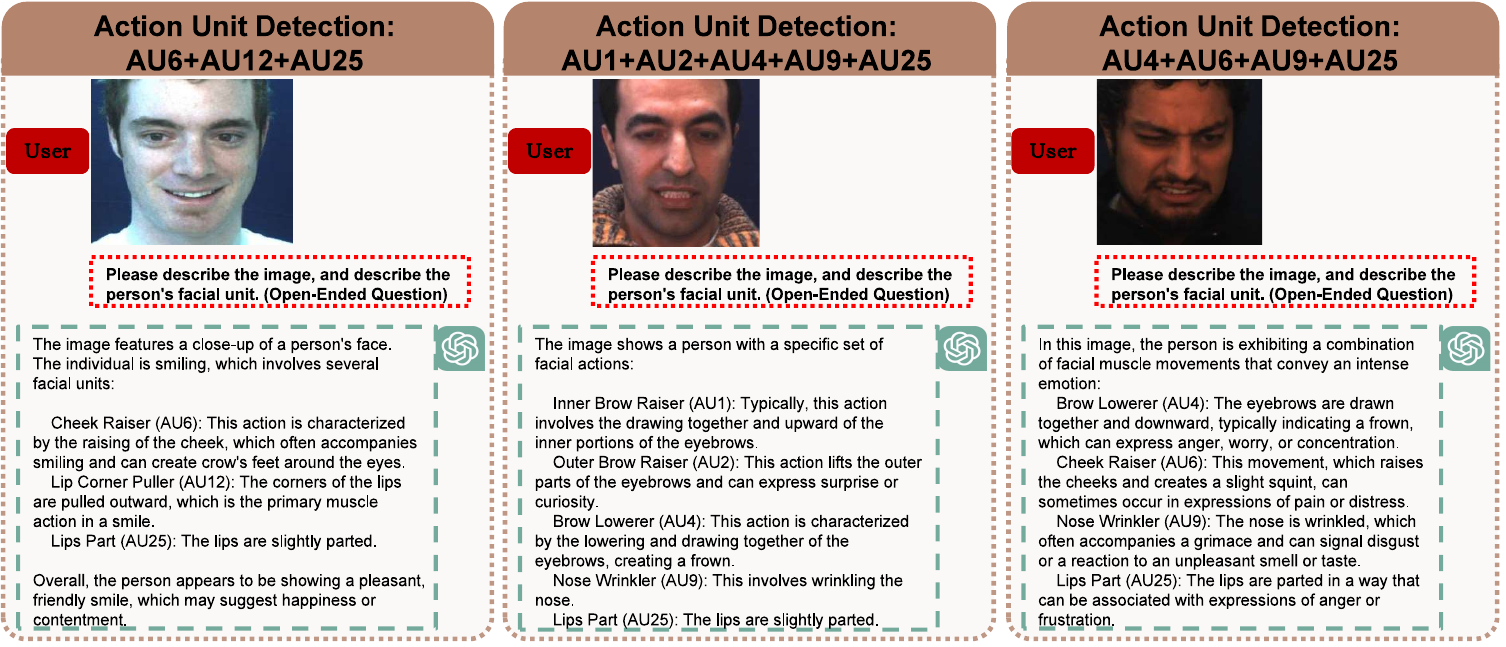}
\end{center}
\vspace{-4mm}
\caption{Action Unit detection on DISFA~\cite{disfa} dataset. We use the single round for the action unit. \gpt can accurately identify each AU.}
\vspace{-4mm}
\label{fig:au}
\end{figure*}

\begin{table*}[ht] 
\centering 
\resizebox{\linewidth}{!}{
\begin{tabular}{cccccccccccccc} 
\toprule
 AU & DRML~\cite{DRML} & DSIN~\cite{corneanu2018deep} & LP~\cite{niu2019local} & SRERL~\cite{li2019semantic} & EAC~\cite{li2018eac} & JAA~\cite{shao2018deep} & ARL~\cite{shao2019facial} & FAUDT~\cite{jacob2021facial} & PIAP~\cite{PIAP-DF} & ME-GraphAU~\cite{ME-graphAU} & BG-AU~\cite{BG-AU} & MPSCL~\cite{MPSCL} & \gpt \\ 
\midrule
 1 & 17.3 & 42.4 & 29.9 & 45.7 & 41.5 & 43.7 & 43.9 & 46.1 & 50.2 & 52.5 & 41.5 & \textbf{62.0} & 52.6   \\ 
 2 & 17.7 & 39.0 & 24.7 & 47.8 & 26.4 & 46.2 & 42.1 & 48.6 & 51.8 & 45.7 & 44.9 & \textbf{65.7} & 56.4   \\ 
 4 & 37.4 & 68.4 & 72.7 & 59.6 & 66.4 & 56.0 & 63.6 & 72.8 & 71.9 & 76.1 & 60.3 & 74.5 & \textbf{82.9}  \\ 
 6 & 29.0 & 28.6 & 46.8 & 47.1 & 50.7 & 41.4 & 41.8 & 56.7 & 50.6 & 51.8 & 51.5 & 53.2 &\textbf{ 64.3}  \\ 
 9 & 10.7 & 46.8 & 49.6 & 45.6 & \textbf{80.5} & 44.7 & 40.0 & 50.0 & 54.5 & 46.5 & 50.3 & 43.1 & 55.3 \\ 
 12 & 37.7 & 70.8 & 72.9 & 73.5 & \textbf{89.3} & 69.6 & 76.2 & 72.1 & 79.7 & 76.1 & 70.4 & 76.9 & 75.4  \\ 
 25 & 38.5 & 90.4 & 93.8 & 84.3 & 88.9 & 88.3 & \textbf{95.2 }& 90.8 & 94.1 & 92.9 & 91.3 & 95.6 & 91.2  \\ 
 26 & 20.1 & 42.2 &65.0	&43.6  &15.6  &58.4	 &\textbf{66.8}  &55.4  & 57.2 & 57.6 & 55.3 & 53.1 & 66.4 \\
\midrule
  Avg. & 26.7 & 53.6 & 56.9 & 55.9 & 48.5 & 56.0 & 58.7 & 61.5 & 63.8 & 62.4 & 58.2 & 65.5 & \textbf{67.3} \\ 
\bottomrule
\end{tabular}}
\caption{Comparison with state-of-the-art methods for AU detection on DISFA~\cite{disfa} dataset using the F1-score metric (in \%).}
\label{tab:au}
\end{table*}

\subsection{Expression Recognition}

\begin{figure*}[!ht]
\begin{center}
\includegraphics[scale=0.7]{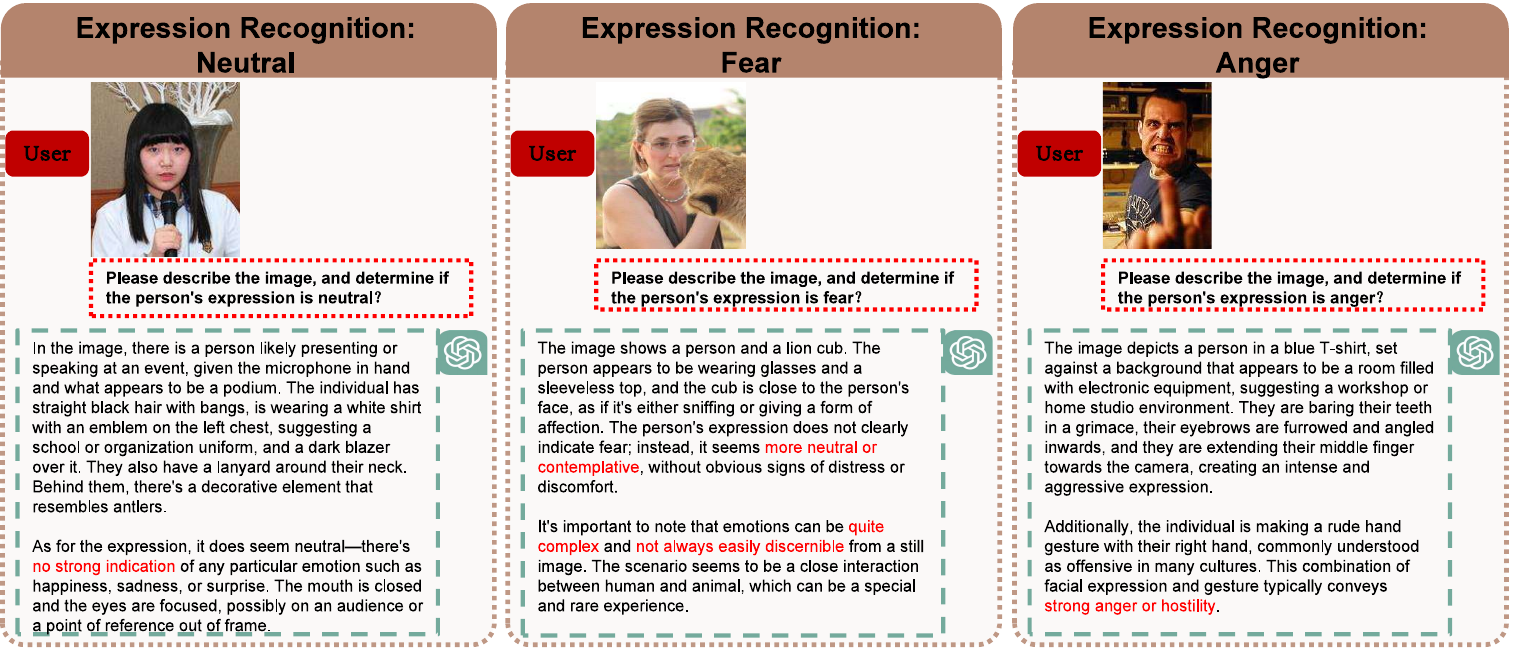}
\end{center}
\vspace{-4mm}
\caption{Expression recognition on RAF-DB~\cite{shan2018reliable} dataset. \gpt cannot achieve good performance on the subjective task of emotion recognition.}
\vspace{-8mm}
\label{fig:er}
\end{figure*}

The facial expression recognition~\cite{shan2018reliable} task involves identifying and analyzing human facial expressions to determine emotions. This task plays a crucial role in understanding human emotions, enhancing communication, and improving mental health monitoring and virtual human companion systems. It can be challenging due to the complexity and variety of facial expressions, as well as the need to detect and classify subtle and transient expressions accurately. For this reason, we qualitatively analyze the performance of \gpt for emotion recognition on RAF-DB~\cite{shan2018reliable} dataset. Our methodology encompassed a multifaceted approach, employing iterative dialogues, open-ended questions, multiple-choice queries, and true/false assessments, specifically utilizing the CASME2 dataset as a basis for evaluation. Contrary to expectations, preliminary results indicate that \gpt exhibits limitations in accurately responding to even basic true/false questions related to emotion recognition, as depicted in the referenced figure shown in Fig.~\ref{fig:er}.

As shown in Fig.~\ref{fig:er}, natural emotions are thought to have no obvious characteristics, as soon as we pass a form of judgment question. For the emotion of Fear, \gpt thinks that the emotion is natural and cannot give the decision of fear. This is because emotions are inherently difficult to recognize without context, which is not considered an objective task. Therefore, \gpt cannot achieve good performance on these subjective tasks. This finding highlights a significant limitation in the application of advanced language models like \gpt for the nuanced task of emotion recognition. It suggests that while such models possess remarkable capabilities in various domains of natural language processing, their effectiveness in interpreting human emotions through facial expressions, especially in the absence of contextual information, remains constrained. The subjective nature of emotional expression, coupled with the subtleties and variations inherent in human facial expressions, necessitates a more sophisticated approach that incorporates contextual understanding and perhaps multimodal inputs that extend beyond textual analysis.

\subsection{Compound Emotion Recognition}
The task of compound emotion recognition~\cite{du2014compound} extends beyond the scope of simple emotion recognition by necessitating the identification and analysis of multiple emotions simultaneously exhibited through human facial expressions. The complexity of this task is amplified by the requirement to accurately detect and classify a spectrum of emotions, which may often be overlapping or present ambiguous signals. It can be more challenging than simple emotion recognition due to the need to detect and classify multiple emotions accurately, as well as the potential for conflicting or ambiguous expressions. In our continued exploration of \gpt’s capabilities, we extend our assessment to include the recognition of compound emotions. 

\begin{figure*}[!ht]
\begin{center}
\includegraphics[scale=0.7]{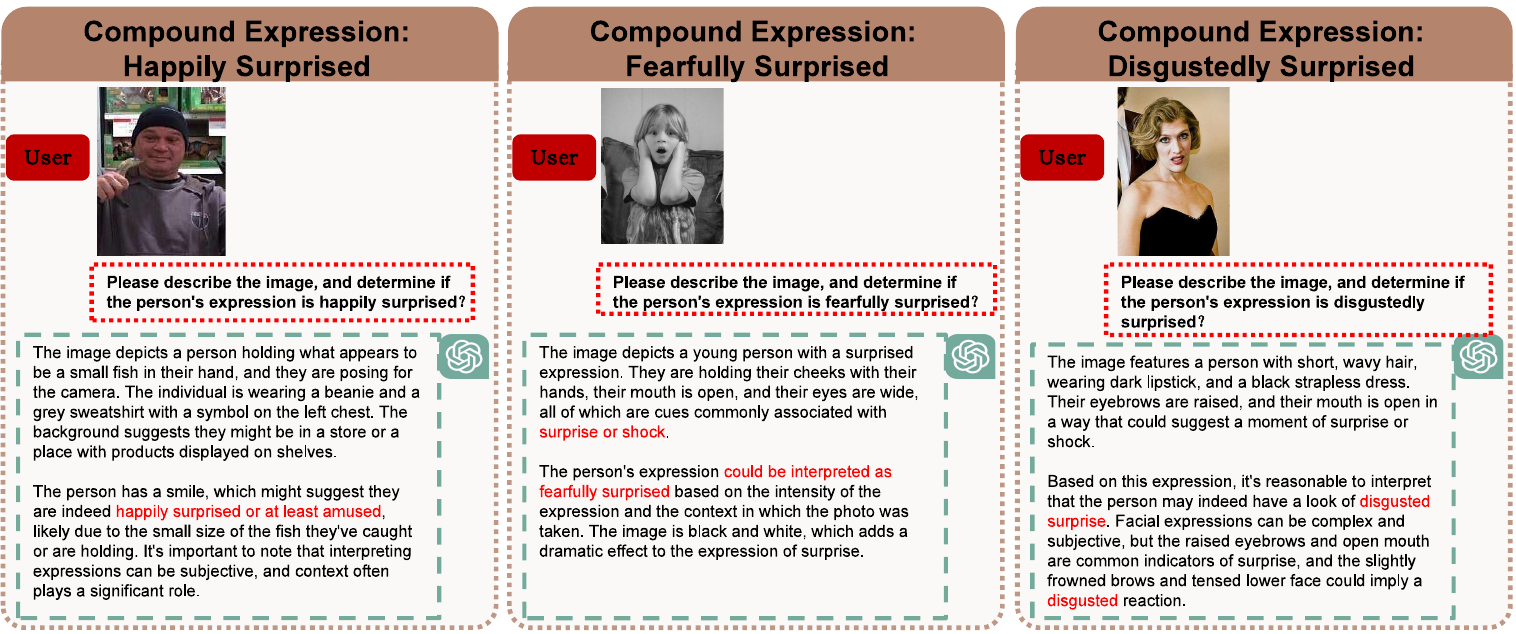}
\vspace{-4mm}
\caption{Compound emotion recognition on RAF-DB~\cite{shan2018reliable} dataset. \gpt can deduce objective compound expressions based on contextual information.}
\label{fig:cer}
\vspace{-8mm}
\end{center}
\end{figure*}

As shown in Fig.~\ref{fig:cer}, we qualitatively analyze the performance of \gpt for compound emotion recognition on RAF-DB~\cite{shan2018reliable} dataset and find that compound expressions can even be recognized. Even compound expressions are recognized more accurately than individual expressions. This does not mean \gpt is more accurate for compound than individual expressions. Instead, this is because the data of this compound expression is relatively more objective, and \gpt has an accurate judgment of this objective expression. This revelation underscores the importance of developing computational models that can navigate the intricacies of human emotions with a high degree of sensitivity and accuracy. For applications in mental health monitoring and virtual companionship, paving the way for innovations in emotional AI that can more closely mimic human empathetic and cognitive processes.

\subsection{Micro-expression Recognition}
The domain of micro-expression~\cite{zhao2023facial} research within emotion recognition is characterized by the endeavor to identify and interpret subtle, fleeting expressions that manifest on the human face. These micro-expressions, often resulting from rapid emotional shifts or attempts to conceal emotions, are particularly ephemeral, lasting only between 1/25 to 1/5 of a second. This attribute renders micro-expressions both a fascinating and formidable area of study~\cite{ASP,SCA,LBP-SIP}. However, the transient and elusive nature of micro-expressions presents significant challenges, notably in their detection and accurate interpretation. In our investigation, we meticulously crafted cue words and deployed a series of experimental setups involving judgment questions, multiple-choice inquiries, and iterative dialogues, all facilitated based on the CASME2 dataset~\cite{CASME} through the \gpt platform. This approach aimed to explore the potential of \gpt in recognizing and interpreting micro-expressions within the constraints of textual communication.

\begin{figure}[!ht]
\begin{center}
\includegraphics[scale=0.6]{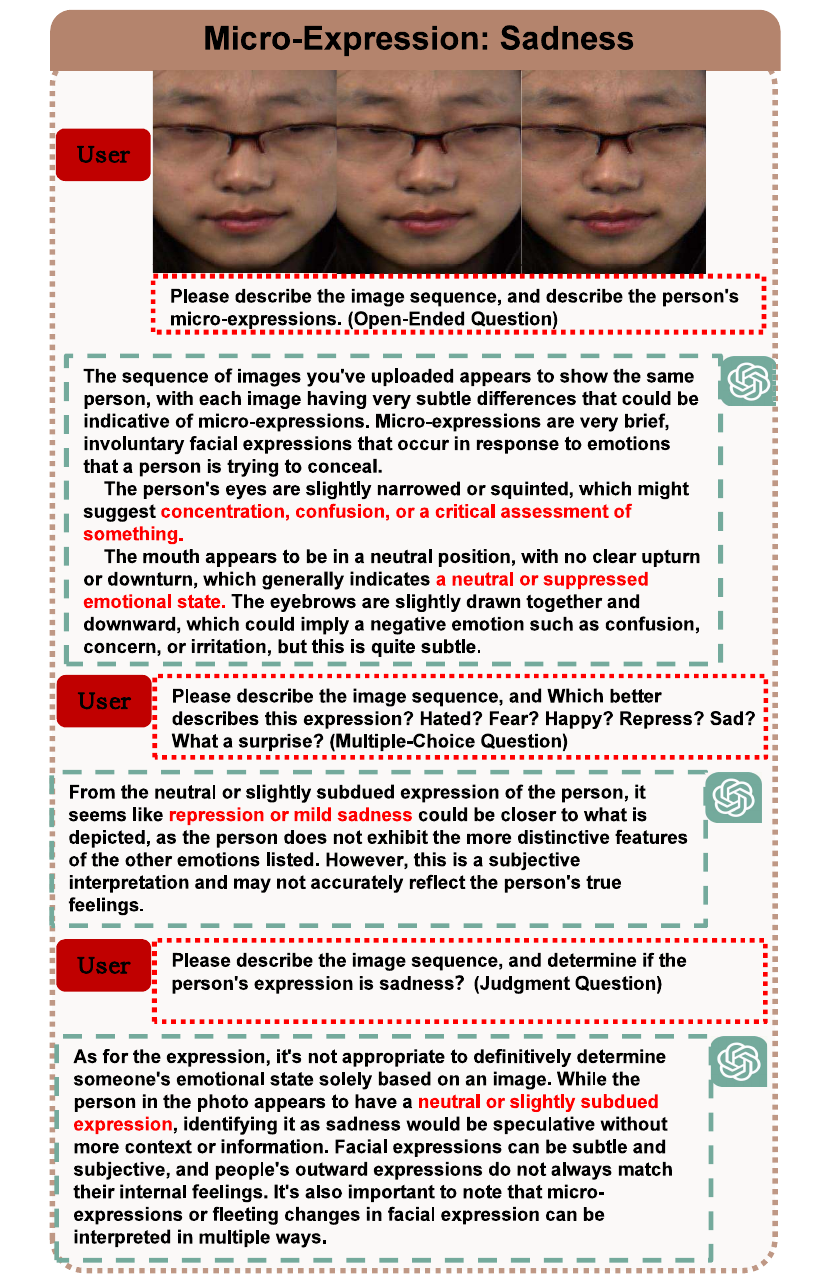}
\end{center}
\caption{Micro-expression recognition on the CASME2~\cite{CASME} dataset. \gpt has difficulty understanding the small differences in the image directly, so it is difficult to understand the micro facial expressions accurately.}
\vspace{-4mm}
\label{fig:me}
\end{figure}

As shown in Fig.~\ref{fig:me}, \gpt did not answer the provided micro-expression test samples satisfactorily. \gpt cannot understand the difference between frames, and the difference is not visible to the human eye. We tried to amplify this difference, but \gpt thought the enlarged image was blurry, so \gpt was very weak on the microexpression task.

\subsection{Micro-gesture Recognition}

Micro-gesture recognition~\cite{liu2021imigue} tasks focus on recognizing and analyzing small, imperceptible body movements and facial expressions produced by people in specific scenarios, which usually represent an individual's inner emotions, attitudes, or cognitive responses. Micro-gesture recognition techniques are valuable for many applications, such as emotion recognition, negotiation, police interrogation, and mental health assessment. The core challenge of this technology is to capture brief and subtle changes in movements that are difficult for individuals to control due to their association with the autonomic nervous system. Micro-gesture recognition improves social interactions and communication by helping people better understand others' emotions and motivations. We carefully designed the cue words and tested several different micro-gesture sequences on judgment questions, multiple-choice questions, and multi-round conversations using \gpt. 

\begin{figure}[!ht]
\begin{center}
\includegraphics[scale=0.6]{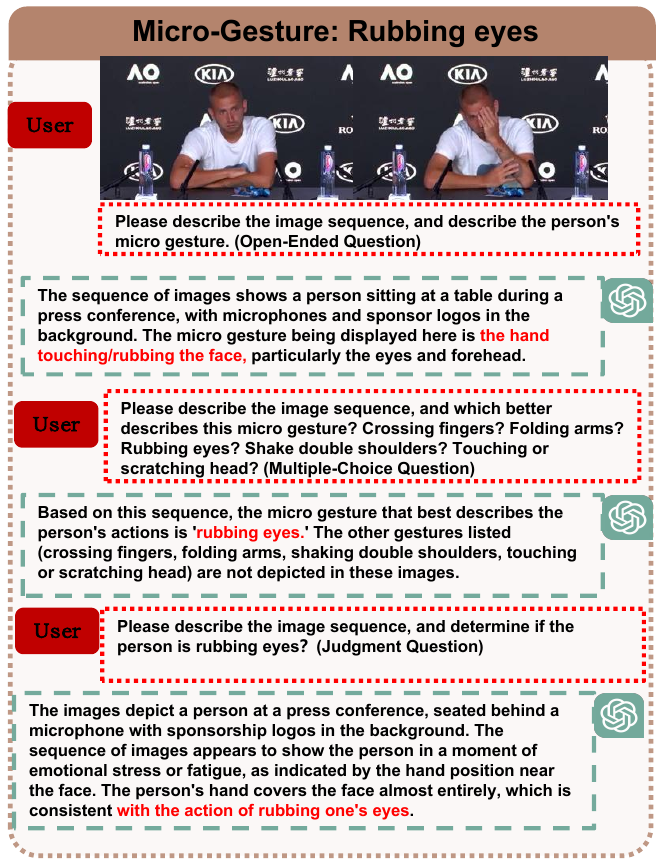}
\end{center}
\caption{Micro-gesture (rubbing eyes) recognition on iMiGUE~\cite{liu2021imigue} dataset. \gpt can recognize obvious movements.}
\label{fig:mg}
\vspace{-8mm}
\end{figure}

\begin{figure}[!ht]
\begin{center}
\includegraphics[scale=0.6]{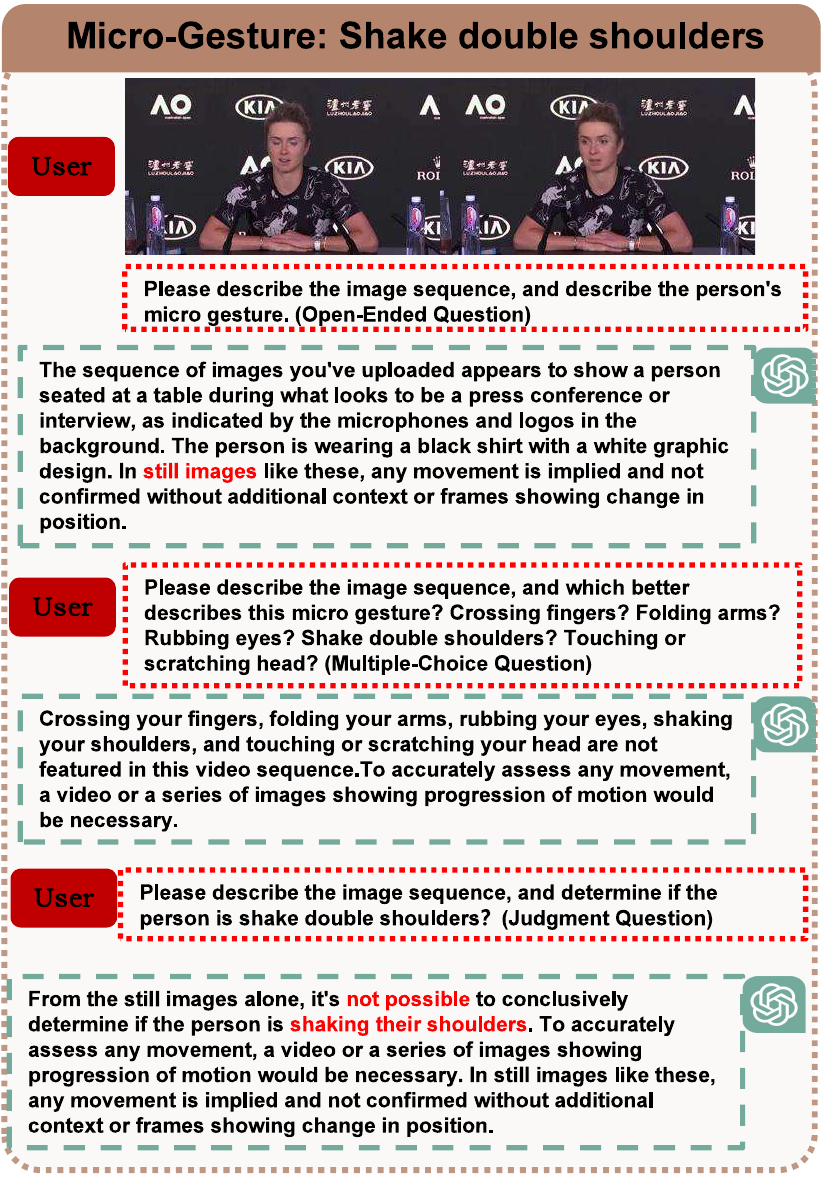}
\end{center}
\vspace{-4mm}
\caption{Micro-gesture (shaking double shoulders) recognition on iMiGUE~\cite{liu2021imigue} dataset. \gpt cannot recognize subtle movements.}
\vspace{-8mm}
\label{fig:mg1}
\end{figure}

We qualitatively analyze the performance of \gpt for micro-gesture recognition on iMiGUE~\cite{liu2021imigue} dataset. As shown in Fig.~\ref{fig:mg}, \gpt can give satisfactory answers to the micro-gesture test samples provided. It can give similar answers to even the most difficult questions (open-ended questions), such as rubbing the face (rubbing the eyes). As shown in Fig.~\ref{fig:mg1}, \gpt doesn't recognize the shoulder flutter. \gpt can't recognize tiny movements. While \gpt marks a significant step forward in the application of AI in the field of emotion and behavior recognition, its current limitations in recognizing certain micro-gestures suggest that further refinement and development is needed.

\subsection{Deception Detection}
Deception detection is an important task for determining the authenticity of video content, which is very important for security. To verify the performance of \gpt for deception detection, we evaluated on Real-Life Trial dataset ~\cite{csen2020multimodal}.

\begin{figure}[!ht]
\begin{center}
\includegraphics[scale=0.8]{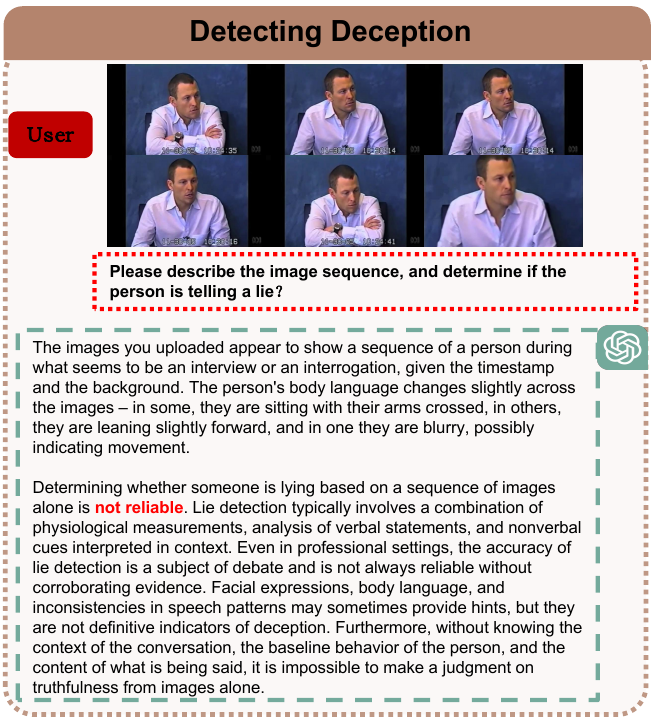}
\end{center}
\caption{Deception Detection on  Real-Life Trial dataset~\cite{csen2020multimodal}. 
\gpt can't recognize a lie or not.}
\label{fig:dc}
\vspace{-8mm}
\end{figure}

As shown in~\ref{fig:dc}, \gpt can't tell if a person in a video is lying. In fact, such subjective tasks are difficult for even real people to accurately judge. In addition, we try to input some multimodal information such as the sound spectrum to guide the \gpt to produce the correct result. But such operations do not allow GPT to reason the correct result. This shows that \gpt is still challenging for subjective tasks.

\vspace{-2mm}

\section{Advanced Capability of Reasoning}

\subsection{Chain of thought}

The concept of Chain-of-Thought (CoT)~\cite{wei2022chain,wang2022self,feng2024towards} was first introduced in the seminal work by researchers at Google, titled Chain-of-Thought Prompting Elicits Reasoning in Large Language Models. This innovative approach represents a significant advancement in cue strategies designed to enhance the performance of Large Language Models (LLMs) in executing complex reasoning tasks, encompassing arithmetic, common sense, and symbolic reasoning domains. Contrary to the Implicit Context Learning (ICL) approach, which primarily relies on input-output pairings, CoT incorporates a series of intermediate inference steps that scaffold toward the final solution, thereby enriching the model's reasoning pathway. In essence, CoT facilitates discrete prompt learning by appending an example to the beginning of a given input, enabling the model to process these concatenated texts simultaneously and produce the desired output. This method, under including additional intermediate prompts for inference, represents a substantial improvement over traditional context learning approaches.\gpt has a hard time recognizing specific expressions without context. However, we asked \gpt to first recognize the specific AU representation and then deduce the emotion based on the relationship between AU and the expression, which allowed \gpt to give some possible outcomes of the expression.

\begin{figure}[t]
\begin{center}
\includegraphics[scale=0.75]{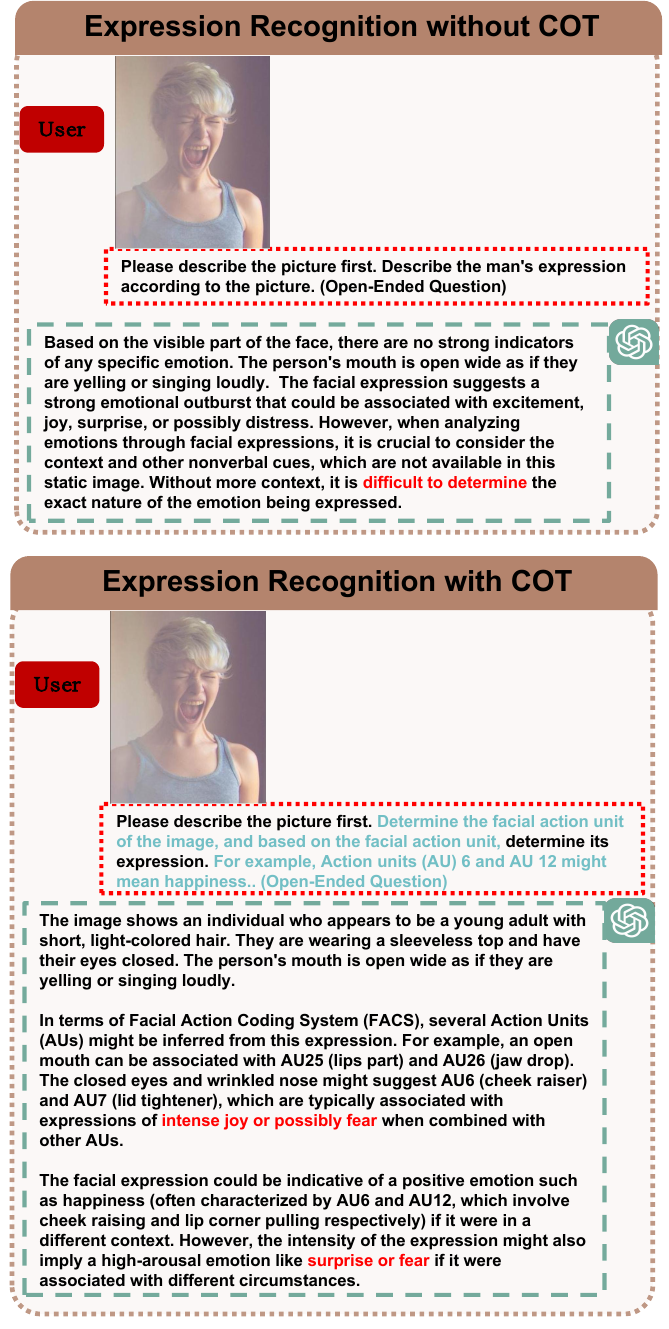}
\end{center}
\vspace{-4mm}
\caption{Improve the accuracy of expression recognition through COT. We use more accurate AU recognition and make \gpt recognition more accurate according to the relationship between AU and expression.}
\vspace{-4mm}
\label{fig:cot}
\end{figure}

Furthermore, the application of CoT in emotion recognition tasks reveals its potential to circumvent some of the limitations faced by models such as \gpt in interpreting ambiguous or neutral expressions. Despite \gpt's proficiency in Action Unit (AU) recognition, its performance in emotion recognition from expressions remains suboptimal. By leveraging the correlation between CoT, AU, and facial expressions, we aim to enhance \gpt's accuracy in this area. As evidenced in Fig.~\ref{fig:cot}, the incorporation of CoT significantly improves \gpt's capability to discern emotions, particularly in instances where expressions are ambiguous or lack clear contextual cues. This methodology enables \gpt to first accurately identify AUs, and subsequently infer the probable emotion based on the established relationship between AUs and facial expressions. The integration of CoT, as illustrated by the blue segments in the figure, thus facilitates a more nuanced understanding and recognition of emotional states by the model. Thus, the application of CoT in affective computing holds the potential to significantly improve the capability of visual language models in interpreting and predicting emotional states with greater accuracy, leveraging contextual information to bridge the gap between task-related cues and the corresponding emotional expressions.

\subsection{Tool call and processing}

\gpt is one of the state-of-the-art multimodal language models that has achieved remarkable success in various natural language processing tasks. However, it is not directly applicable to some complex tasks, such as remote photoplethysmography (rPPG)~\cite{huang2023challenges,dual-GAN,wang2016algorithmic,yu2021facial,yu2022physformer,yu2023physformer,lu2023neuron,niu2019rhythmnet,wang2015novel,chen2018deepphys,liu2024rppg}. rPPG is a non-invasive technique used to measure heart rate and respiratory rate from facial videos. It has a wide range of applications in healthcare, entertainment, and human-computer interaction. 

\begin{figure}[!ht]
\begin{center}
\includegraphics[scale=0.6]{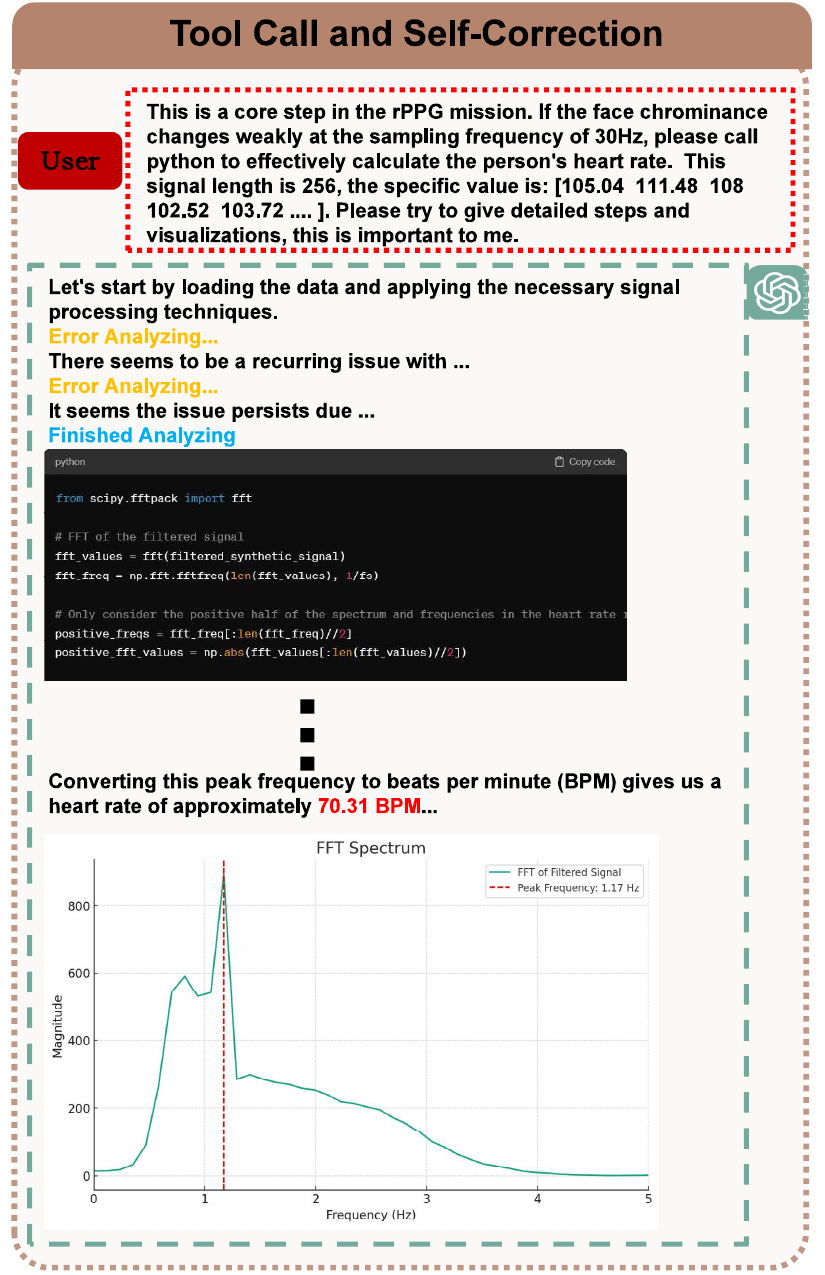}
\end{center}
\caption{Tool call and processing for rPPG task. \gpt can directly write programs for signal processing on request and try to run them. If there is an error, \gpt can be further corrected according to the error prompt, and finally give the heart rate prediction and visualization.}
\label{fig:au}
\vspace{-4mm}
\end{figure}

Unfortunately, \gpt cannot read in long time-series videos and cannot discern subtle chromatic variations. To address this issue, a solution is for professional researchers to collaborate with \gpt. In this regard, we found that \gpt can call Python tools to run code and debug it. To demonstrate this process, we extracted facial video chromatic changes and used \gpt to process the signal. As shown in the figure, \gpt called Python to process and visualize the signal. During this process, there were several bugs, but \gpt was able to self-correct based on the bug information and ultimately provided an accurate heart rate result.

This process has provided us with an insight that this can be turned into a framework for a human-large language multimodal model that can self-correct. In this process, any large model can self-correct. This framework has immense potential in enhancing the accuracy and efficiency of various tasks, including rPPG.

\section{Further Discussion}
\gpt is a powerful language model that has shown remarkable success in various natural language processing tasks. However, it faces several challenges in other domains, such as facial expression recognition, emotion recognition, complex emotion recognition, non-contact physiological measurement, and authenticity detection.

Emotion recognition is the process of identifying and classifying emotions based on physiological signals, facial expressions, and speech patterns. \gpt has shown promising results in this task; however, it requires a large amount of training data and may not generalize well to new datasets. To overcome this limitation, future research can focus on developing transfer learning techniques that enable \gpt to learn from smaller datasets and generalize to new datasets.

Non-contact physiological measurement involves measuring physiological signals, such as heart rate, respiratory rate, and blood pressure, without direct contact with the body~\cite{huang2023challenges,dual-GAN,lu2023neuron,wang2016algorithmic,wang2015novel,chen2018deepphys}. \gpt faces difficulty in this task due to its limited ability to process and interpret physiological signals accurately. To overcome this limitation, future research can focus on developing new technologies that can capture physiological signals accurately and integrate them with \gpt.

Deception detection involves identifying and verifying the authenticity of a person or an object. \gpt faces difficulty in this task due to its limited ability to process and interpret visual and audio information accurately. To overcome this limitation, future research can explore ways to integrate \gpt with computer vision and audio processing techniques to improve authenticity detection accuracy.

In conclusion, \gpt faces several challenges in non-language tasks, such as facial expression recognition, emotion recognition, complex emotion recognition, non-contact physiological measurement, and deception detection. Future research can focus on developing new techniques to enhance \gpt's ability to process and integrate multimodal data and improve its accuracy and efficiency in these tasks.

\section{Conclusion}

In this paper, we have discussed the challenges that \gpt faces in non-language tasks, such as facial expression recognition, emotion recognition, complex emotion recognition, non-contact physiological measurement, and authenticity detection. While \gpt has shown remarkable success in various natural language processing tasks, it faces limitations in these domains due to its limited ability to process and interpret visual and audio information accurately. To overcome these limitations, future research can focus on developing new techniques to enhance \gpt's ability to process and integrate multimodal data and improve its accuracy and efficiency in these tasks. This may involve integrating \gpt with computer vision and audio processing techniques, developing transfer learning techniques, exploring new sensor technologies, and improving the quality and quantity of training data. By addressing these challenges, \gpt has the potential to significantly advance the fields of facial expression recognition, emotion recognition, complex emotion recognition, non-contact physiological measurement, and authenticity detection, and open up new avenues for research and application in these domains.

{
    \small
    \bibliographystyle{ieeenat_fullname}
    \bibliography{main}

\begin{thebibliography}{49}
\providecommand{\natexlab}[1]{#1}
\providecommand{\url}[1]{\texttt{#1}}
\expandafter\ifx\csname urlstyle\endcsname\relax
  \providecommand{\doi}[1]{doi: #1}\else
  \providecommand{\doi}{doi: \begingroup \urlstyle{rm}\Url}\fi

\bibitem[Achiam et~al.(2023)Achiam, Adler, Agarwal, Ahmad, Akkaya, Aleman, Almeida, Altenschmidt, Altman, Anadkat, et~al.]{achiam2023gpt}
Josh Achiam, Steven Adler, Sandhini Agarwal, Lama Ahmad, Ilge Akkaya, Florencia~Leoni Aleman, Diogo Almeida, Janko Altenschmidt, Sam Altman, Shyamal Anadkat, et~al.
\newblock Gpt-4 technical report.
\newblock \emph{arXiv preprint arXiv:2303.08774}, 2023.

\bibitem[Chen and McDuff(2018)]{chen2018deepphys}
Weixuan Chen and Daniel McDuff.
\newblock Deepphys: Video-based physiological measurement using convolutional attention networks.
\newblock In \emph{Proceedings of the european conference on computer vision (ECCV)}, pages 349--365, 2018.

\bibitem[Corneanu et~al.(2018)Corneanu, Madadi, and Escalera]{corneanu2018deep}
Ciprian Corneanu, Meysam Madadi, and Sergio Escalera.
\newblock Deep structure inference network for facial action unit recognition.
\newblock In \emph{Proceedings of the european conference on computer vision (ECCV)}, pages 298--313, 2018.

\bibitem[Cui et~al.(2023)Cui, Kuang, Gao, Talamadupula, and Ji]{BG-AU}
Zijun Cui, Chenyi Kuang, Tian Gao, Kartik Talamadupula, and Qiang Ji.
\newblock Biomechanics-guided facial action unit detection through force modeling.
\newblock In \emph{Proceedings of the IEEE/CVF Conference on Computer Vision and Pattern Recognition}, pages 8694--8703, 2023.

\bibitem[Du et~al.(2014)Du, Tao, and Martinez]{du2014compound}
Shichuan Du, Yong Tao, and Aleix~M Martinez.
\newblock Compound facial expressions of emotion.
\newblock \emph{Proceedings of the national academy of sciences}, 111\penalty0 (15):\penalty0 E1454--E1462, 2014.

\bibitem[Ekman and Friesen(1978)]{FACS}
Paul Ekman and Wallace~V Friesen.
\newblock Facial action coding system.
\newblock \emph{Environmental Psychology \& Nonverbal Behavior}, 1978.

\bibitem[Feng et~al.(2024)Feng, Zhang, Gu, Ye, He, and Wang]{feng2024towards}
Guhao Feng, Bohang Zhang, Yuntian Gu, Haotian Ye, Di He, and Liwei Wang.
\newblock Towards revealing the mystery behind chain of thought: a theoretical perspective.
\newblock \emph{Advances in Neural Information Processing Systems}, 36, 2024.

\bibitem[Guo et~al.(2023)Guo, Selvaraj, Yu, Kong, Shen, and Kot]{guo2023audio}
Xiaobao Guo, Nithish~Muthuchamy Selvaraj, Zitong Yu, Adams Wai-Kin Kong, Bingquan Shen, and Alex Kot.
\newblock Audio-visual deception detection: Dolos dataset and parameter-efficient crossmodal learning.
\newblock In \emph{Proceedings of the IEEE/CVF International Conference on Computer Vision}, pages 22135--22145, 2023.

\bibitem[Huang et~al.(2023)Huang, Hu, Liu, Lin, Su, Zhao, Wang, and Wang]{huang2023challenges}
Bin Huang, Shen Hu, Zimeng Liu, Chun-Liang Lin, Junfeng Su, Changchen Zhao, Li Wang, and Wenjin Wang.
\newblock Challenges and prospects of visual contactless physiological monitoring in clinical study.
\newblock \emph{NPJ Digital Medicine}, 6\penalty0 (1):\penalty0 231, 2023.

\bibitem[Jacob and Stenger(2021)]{jacob2021facial}
Geethu~Miriam Jacob and Bjorn Stenger.
\newblock Facial action unit detection with transformers.
\newblock In \emph{Proceedings of the IEEE/CVF Conference on Computer Vision and Pattern Recognition}, pages 7680--7689, 2021.

\bibitem[Li et~al.(2019)Li, Zhu, Zeng, Wang, and Lin]{li2019semantic}
Guanbin Li, Xin Zhu, Yirui Zeng, Qing Wang, and Liang Lin.
\newblock Semantic relationships guided representation learning for facial action unit recognition.
\newblock In \emph{Proceedings of the AAAI Conference on Artificial Intelligence}, pages 8594--8601, 2019.

\bibitem[Li et~al.(2023)Li, He, Wang, Li, Wang, Luo, Wang, Wang, and Qiao]{li2023videochat}
KunChang Li, Yinan He, Yi Wang, Yizhuo Li, Wenhai Wang, Ping Luo, Yali Wang, Limin Wang, and Yu Qiao.
\newblock Videochat: Chat-centric video understanding.
\newblock \emph{arXiv preprint arXiv:2305.06355}, 2023.

\bibitem[Li et~al.(2018)Li, Abtahi, Zhu, and Yin]{li2018eac}
Wei Li, Farnaz Abtahi, Zhigang Zhu, and Lijun Yin.
\newblock Eac-net: Deep nets with enhancing and cropping for facial action unit detection.
\newblock \emph{IEEE transactions on pattern analysis and machine intelligence}, 40\penalty0 (11):\penalty0 2583--2596, 2018.

\bibitem[Li et~al.(2021{\natexlab{a}})Li, Huang, and Zhao]{SCA}
Yante Li, Xiaohua Huang, and Guoying Zhao.
\newblock Micro-expression action unit detection with spatial and channel attention.
\newblock \emph{Neurocomputing}, 436:\penalty0 221--231, 2021{\natexlab{a}}.

\bibitem[Li et~al.(2021{\natexlab{b}})Li, Peng, and Zhao]{ASP}
Yante Li, Wei Peng, and Guoying Zhao.
\newblock Micro-expression action unit detection with dual-view attentive similarity-preserving knowledge distillation.
\newblock In \emph{2021 16th IEEE International Conference on Automatic Face and Gesture Recognition (FG 2021)}, pages 01--08. IEEE, 2021{\natexlab{b}}.

\bibitem[Liu et~al.(2021)Liu, Shi, Chen, Yu, Li, and Zhao]{liu2021imigue}
Xin Liu, Henglin Shi, Haoyu Chen, Zitong Yu, Xiaobai Li, and Guoying Zhao.
\newblock imigue: An identity-free video dataset for micro-gesture understanding and emotion analysis.
\newblock In \emph{Proceedings of the IEEE/CVF conference on computer vision and pattern recognition}, pages 10631--10642, 2021.

\bibitem[Liu et~al.(2023)Liu, Yuan, Niu, Shi, Yu, Yue, and Yang]{MPSCL}
Xin Liu, Kaishen Yuan, Xuesong Niu, Jingang Shi, Zitong Yu, Huanjing Yue, and Jingyu Yang.
\newblock Multi-scale promoted self-adjusting correlation learning for facial action unit detection.
\newblock \emph{arXiv preprint arXiv:2308.07770}, 2023.

\bibitem[Liu et~al.(2024)Liu, Zhang, Yu, Lu, Yue, and Yang]{liu2024rppg}
Xin Liu, Yuting Zhang, Zitong Yu, Hao Lu, Huanjing Yue, and Jingyu Yang.
\newblock rppg-mae: Self-supervised pretraining with masked autoencoders for remote physiological measurements.
\newblock \emph{IEEE Transactions on Multimedia}, 2024.

\bibitem[Lu et~al.(2021)Lu, Han, and Zhou]{dual-GAN}
Hao Lu, Hu Han, and S~Kevin Zhou.
\newblock Dual-gan: Joint bvp and noise modeling for remote physiological measurement.
\newblock In \emph{Proceedings of the IEEE/CVF conference on computer vision and pattern recognition}, pages 12404--12413, 2021.

\bibitem[Lu et~al.(2023)Lu, Yu, Niu, and Chen]{lu2023neuron}
Hao Lu, Zitong Yu, Xuesong Niu, and Ying-Cong Chen.
\newblock Neuron structure modeling for generalizable remote physiological measurement.
\newblock In \emph{Proceedings of the IEEE/CVF Conference on Computer Vision and Pattern Recognition}, pages 18589--18599, 2023.

\bibitem[Luo et~al.(2022)Luo, Song, Xie, Shen, and Gunes]{ME-graphAU}
Cheng Luo, Siyang Song, Weicheng Xie, Linlin Shen, and Hatice Gunes.
\newblock Learning multi-dimensional edge feature-based au relation graph for facial action unit recognition.
\newblock \emph{international joint conference on artificial intelligence}, 2022.

\bibitem[Luo et~al.(2023)Luo, Zhao, Yang, Dong, Qiu, Lu, Wang, and Wei]{luo2023valley}
Ruipu Luo, Ziwang Zhao, Min Yang, Junwei Dong, Minghui Qiu, Pengcheng Lu, Tao Wang, and Zhongyu Wei.
\newblock Valley: Video assistant with large language model enhanced ability.
\newblock \emph{arXiv preprint arXiv:2306.07207}, 2023.

\bibitem[Lv and Sun(2024)]{lv2024video}
Hui Lv and Qianru Sun.
\newblock Video anomaly detection and explanation via large language models.
\newblock \emph{arXiv preprint arXiv:2401.05702}, 2024.

\bibitem[Maaz et~al.(2023)Maaz, Rasheed, Khan, and Khan]{maaz2023video}
Muhammad Maaz, Hanoona Rasheed, Salman Khan, and Fahad~Shahbaz Khan.
\newblock Video-chatgpt: Towards detailed video understanding via large vision and language models.
\newblock \emph{arXiv preprint arXiv:2306.05424}, 2023.

\bibitem[Mavadati et~al.(2013)Mavadati, Mahoor, Bartlett, Trinh, and Cohn]{disfa}
S~Mohammad Mavadati, Mohammad~H Mahoor, Kevin Bartlett, Philip Trinh, and Jeffrey~F Cohn.
\newblock Disfa: A spontaneous facial action intensity database.
\newblock \emph{IEEE Transactions on Affective Computing}, 4\penalty0 (2):\penalty0 151--160, 2013.

\bibitem[Niu et~al.(2019{\natexlab{a}})Niu, Han, Yang, Huang, and Shan]{niu2019local}
Xuesong Niu, Hu Han, Songfan Yang, Yan Huang, and Shiguang Shan.
\newblock Local relationship learning with person-specific shape regularization for facial action unit detection.
\newblock In \emph{Proceedings of the IEEE/CVF Conference on computer vision and pattern recognition}, pages 11917--11926, 2019{\natexlab{a}}.

\bibitem[Niu et~al.(2019{\natexlab{b}})Niu, Shan, Han, and Chen]{niu2019rhythmnet}
Xuesong Niu, Shiguang Shan, Hu Han, and Xilin Chen.
\newblock Rhythmnet: End-to-end heart rate estimation from face via spatial-temporal representation.
\newblock \emph{IEEE Transactions on Image Processing}, 29:\penalty0 2409--2423, 2019{\natexlab{b}}.

\bibitem[Poria et~al.(2017)Poria, Cambria, Bajpai, and Hussain]{poria2017review}
Soujanya Poria, Erik Cambria, Rajiv Bajpai, and Amir Hussain.
\newblock A review of affective computing: From unimodal analysis to multimodal fusion.
\newblock \emph{Information fusion}, 37:\penalty0 98--125, 2017.

\bibitem[{\c{S}}en et~al.(2020){\c{S}}en, Perez-Rosas, Yanikoglu, Abouelenien, Burzo, and Mihalcea]{csen2020multimodal}
M~Umut {\c{S}}en, Veronica Perez-Rosas, Berrin Yanikoglu, Mohamed Abouelenien, Mihai Burzo, and Rada Mihalcea.
\newblock Multimodal deception detection using real-life trial data.
\newblock \emph{IEEE Transactions on Affective Computing}, 13\penalty0 (1):\penalty0 306--319, 2020.

\bibitem[Shan and Deng(2018)]{shan2018reliable}
Li Shan and Weihong Deng.
\newblock Reliable crowdsourcing and deep locality-preserving learning for unconstrained facial expression recognition.
\newblock \emph{IEEE Transactions on Image Processing}, 28\penalty0 (1):\penalty0 356--370, 2018.

\bibitem[Shao et~al.(2018)Shao, Liu, Cai, and Ma]{shao2018deep}
Zhiwen Shao, Zhilei Liu, Jianfei Cai, and Lizhuang Ma.
\newblock Deep adaptive attention for joint facial action unit detection and face alignment.
\newblock In \emph{Proceedings of the European conference on computer vision (ECCV)}, pages 705--720, 2018.

\bibitem[Shao et~al.(2019)Shao, Liu, Cai, Wu, and Ma]{shao2019facial}
Zhiwen Shao, Zhilei Liu, Jianfei Cai, Yunsheng Wu, and Lizhuang Ma.
\newblock Facial action unit detection using attention and relation learning.
\newblock \emph{IEEE transactions on affective computing}, 13\penalty0 (3):\penalty0 1274--1289, 2019.

\bibitem[Tang et~al.(2021)Tang, Zeng, Zhao, and Zhang]{PIAP-DF}
Yang Tang, Wangding Zeng, Dafei Zhao, and Honggang Zhang.
\newblock Piap-df: Pixel-interested and anti person-specific facial action unit detection net with discrete feedback learning.
\newblock In \emph{Proceedings of the IEEE/CVF International Conference on Computer Vision}, pages 12899--12908, 2021.

\bibitem[Wang et~al.(2022{\natexlab{a}})Wang, Ahn, and Kim]{wang2022self}
Hao Wang, Euijoon Ahn, and Jinman Kim.
\newblock Self-supervised representation learning framework for remote physiological measurement using spatiotemporal augmentation loss.
\newblock In \emph{Proceedings of the AAAI Conference on Artificial Intelligence}, pages 2431--2439, 2022{\natexlab{a}}.

\bibitem[Wang et~al.(2015{\natexlab{a}})Wang, Stuijk, and De~Haan]{wang2015novel}
Wenjin Wang, Sander Stuijk, and Gerard De~Haan.
\newblock A novel algorithm for remote photoplethysmography: Spatial subspace rotation.
\newblock \emph{IEEE transactions on biomedical engineering}, 63\penalty0 (9):\penalty0 1974--1984, 2015{\natexlab{a}}.

\bibitem[Wang et~al.(2016)Wang, Den~Brinker, Stuijk, and De~Haan]{wang2016algorithmic}
Wenjin Wang, Albertus~C Den~Brinker, Sander Stuijk, and Gerard De~Haan.
\newblock Algorithmic principles of remote ppg.
\newblock \emph{IEEE Transactions on Biomedical Engineering}, 64\penalty0 (7):\penalty0 1479--1491, 2016.

\bibitem[Wang et~al.(2015{\natexlab{b}})Wang, See, Phan, and Oh]{LBP-SIP}
Yandan Wang, John See, Raphael C-W Phan, and Yee-Hui Oh.
\newblock Lbp with six intersection points: Reducing redundant information in lbp-top for micro-expression recognition.
\newblock In \emph{Computer Vision--ACCV 2014: 12th Asian Conference on Computer Vision, Singapore, Singapore, November 1-5, 2014, Revised Selected Papers, Part I 12}, pages 525--537. Springer, 2015{\natexlab{b}}.

\bibitem[Wang et~al.(2022{\natexlab{b}})Wang, Song, Tao, Liotta, Yang, Li, Gao, Sun, Ge, Zhang, et~al.]{wang2022systematic}
Yan Wang, Wei Song, Wei Tao, Antonio Liotta, Dawei Yang, Xinlei Li, Shuyong Gao, Yixuan Sun, Weifeng Ge, Wei Zhang, et~al.
\newblock A systematic review on affective computing: Emotion models, databases, and recent advances.
\newblock \emph{Information Fusion}, 83:\penalty0 19--52, 2022{\natexlab{b}}.

\bibitem[Wei et~al.(2022)Wei, Wang, Schuurmans, Bosma, Xia, Chi, Le, Zhou, et~al.]{wei2022chain}
Jason Wei, Xuezhi Wang, Dale Schuurmans, Maarten Bosma, Fei Xia, Ed Chi, Quoc~V Le, Denny Zhou, et~al.
\newblock Chain-of-thought prompting elicits reasoning in large language models.
\newblock \emph{Advances in Neural Information Processing Systems}, 35:\penalty0 24824--24837, 2022.

\bibitem[Wen et~al.(2023)Wen, Yang, Fu, Wang, Cai, Li, Ma, Li, Xu, Shang, et~al.]{wen2023road}
Licheng Wen, Xuemeng Yang, Daocheng Fu, Xiaofeng Wang, Pinlong Cai, Xin Li, Tao Ma, Yingxuan Li, Linran Xu, Dengke Shang, et~al.
\newblock On the road with gpt-4v (ision): Early explorations of visual-language model on autonomous driving.
\newblock \emph{arXiv preprint arXiv:2311.05332}, 2023.

\bibitem[Wilie et~al.(2020)Wilie, Vincentio, Winata, Cahyawijaya, Li, Lim, Soleman, Mahendra, Fung, Bahar, et~al.]{wilie2020indonlu}
Bryan Wilie, Karissa Vincentio, Genta~Indra Winata, Samuel Cahyawijaya, Xiaohong Li, Zhi~Yuan Lim, Sidik Soleman, Rahmad Mahendra, Pascale Fung, Syafri Bahar, et~al.
\newblock Indonlu: Benchmark and resources for evaluating indonesian natural language understanding.
\newblock \emph{arXiv preprint arXiv:2009.05387}, 2020.

\bibitem[Xu et~al.(2020)Xu, Hu, Zhang, Li, Cao, Li, Xu, Sun, Yu, Yu, et~al.]{xu2020clue}
Liang Xu, Hai Hu, Xuanwei Zhang, Lu Li, Chenjie Cao, Yudong Li, Yechen Xu, Kai Sun, Dian Yu, Cong Yu, et~al.
\newblock Clue: A chinese language understanding evaluation benchmark.
\newblock \emph{arXiv preprint arXiv:2004.05986}, 2020.

\bibitem[Yan et~al.(2014)Yan, Li, Wang, Zhao, Liu, Chen, and Fu]{CASME}
Wen-Jing Yan, Xiaobai Li, Su-Jing Wang, Guoying Zhao, Yong-Jin Liu, Yu-Hsin Chen, and Xiaolan Fu.
\newblock Casme ii: An improved spontaneous micro-expression database and the baseline evaluation.
\newblock \emph{PloS one}, 9\penalty0 (1):\penalty0 e86041, 2014.

\bibitem[Ye et~al.(2023)Ye, Xu, Xu, Ye, Yan, Zhou, Wang, Hu, Shi, Shi, et~al.]{ye2023mplug}
Qinghao Ye, Haiyang Xu, Guohai Xu, Jiabo Ye, Ming Yan, Yiyang Zhou, Junyang Wang, Anwen Hu, Pengcheng Shi, Yaya Shi, et~al.
\newblock mplug-owl: Modularization empowers large language models with multimodality.
\newblock \emph{arXiv preprint arXiv:2304.14178}, 2023.

\bibitem[Yu et~al.(2021)Yu, Li, and Zhao]{yu2021facial}
Zitong Yu, Xiaobai Li, and Guoying Zhao.
\newblock Facial-video-based physiological signal measurement: Recent advances and affective applications.
\newblock \emph{IEEE Signal Processing Magazine}, 38\penalty0 (6):\penalty0 50--58, 2021.

\bibitem[Yu et~al.(2022)Yu, Shen, Shi, Zhao, Torr, and Zhao]{yu2022physformer}
Zitong Yu, Yuming Shen, Jingang Shi, Hengshuang Zhao, Philip~HS Torr, and Guoying Zhao.
\newblock Physformer: Facial video-based physiological measurement with temporal difference transformer.
\newblock In \emph{Proceedings of the IEEE/CVF conference on computer vision and pattern recognition}, pages 4186--4196, 2022.

\bibitem[Yu et~al.(2023)Yu, Shen, Shi, Zhao, Cui, Zhang, Torr, and Zhao]{yu2023physformer}
Zitong Yu, Yuming Shen, Jingang Shi, Hengshuang Zhao, Yawen Cui, Jiehua Zhang, Philip Torr, and Guoying Zhao.
\newblock Physformer++: Facial video-based physiological measurement with slowfast temporal difference transformer.
\newblock \emph{International Journal of Computer Vision}, 131\penalty0 (6):\penalty0 1307--1330, 2023.

\bibitem[Zhao et~al.(2023)Zhao, Li, Li, and Pietik{\"a}inen]{zhao2023facial}
Guoying Zhao, Xiaobai Li, Yante Li, and Matti Pietik{\"a}inen.
\newblock Facial micro-expressions: An overview.
\newblock \emph{Proceedings of the IEEE}, 2023.

\bibitem[Zhao et~al.(2016)Zhao, Chu, and Zhang]{DRML}
Kaili Zhao, Wen-Sheng Chu, and Honggang Zhang.
\newblock Deep region and multi-label learning for facial action unit detection.
\newblock In \emph{Proceedings of the IEEE conference on computer vision and pattern recognition}, pages 3391--3399, 2016.

\end{thebibliography}
}


\end{document}